\pdfoutput=1

\documentclass[11pt]{article}
\usepackage{authblk}
\usepackage{graphicx}
\usepackage[]{ACL2023}

\usepackage{times}
\usepackage{latexsym}
\usepackage{subcaption}
\usepackage{algorithm}
\usepackage{algpseudocode}

\usepackage[T1]{fontenc}

\usepackage[utf8]{inputenc}

\usepackage{microtype}
\usepackage{hyperref}

\usepackage{inconsolata}
\usepackage{algpseudocode}

\title{From the New World of Word Embeddings: A Comparative Study of Small-World Lexico-Semantic Networks in LLMs}

\author[1]{\textbf{Zhu Liu}}
\author[1]{\textbf{Ying Liu}}
\author[2]{\textbf{Kangyang Luo}}
\author[2]{\textbf{Cunliang Kong}}
\author[2]{\textbf{Maosong Sun}}

\affil[1]{School of Humanities, Tsinghua University}
\affil[2]{Department of Computer Science and Technology, Tsinghua University}

\affil[ ]{\nolinkurl{liuzhu22@mails.tsinghua.edu.cn}}

\usepackage{booktabs}
\usepackage{tabularx}
\usepackage{caption}
\usepackage{amsfonts}
\usepackage{amsmath}
\usepackage{color,xcolor} 
\usepackage{xspace}
\usepackage{adjustbox}
\usepackage{multirow}

\usepackage{amssymb}
\usepackage{algpseudocode}
\usepackage{xcolor,colortbl}

\makeatletter

\newcommand{\Rmnum}[1]{\mathrm{\expandafter\@slowromancap\romannumeral #1@}}
\makeatother

\begin{document}

\setcounter{page}{1}

\maketitle

\vspace{2cm} 

\begin{abstract}

Lexico-semantic networks represent words as nodes and their semantic relatedness as edges. While such networks are traditionally constructed using embeddings from encoder-based models or static vectors, embeddings from decoder-only large language models (LLMs) remain underexplored. Unlike encoder models, LLMs are trained with a next-token prediction objective, which does not directly encode the meaning of the current token. In this paper, we construct lexico-semantic networks from the input embeddings of LLMs with varying parameter scales and conduct a comparative analysis of their global and local structures. Our results show that these networks exhibit small-world properties, characterized by high clustering and short path lengths. Moreover, larger LLMs yield more intricate networks with less small-world effects and longer paths, reflecting richer semantic structures and relations. We further validate our approach through analyses of common conceptual pairs, structured lexical relations derived from WordNet, and a cross-lingual semantic network for qualitative words.

\end{abstract}

\section{Introduction}
\label{introduction}

What does the meaning of a word come from? According to the distributional hypothesis~\cite{harris1954distributional,firth1957synopsis,boleda2020distributional}, it arises from the relationships of the word with other words. These relationships - whether syntagmatic (co-occurrence within the same context) or paradigmatic (sharing similar neighboring words) - can be modeled as a network connecting different but related words. This distributional perspective on meaning, along with network-based representations, has been widely explored in areas such as semantic lexicons (e.g., WordNet-inspired resources)~\cite{miller1995wordnet}, language typology (e.g., semantic map models)~\cite{haspelmath2003geometry} and cognitive science (e.g., conceptual space framework)~\cite{gardenfors2000conceptual}.

Lexico-semantic networks can be built manually based on conceptual spaces~\cite{gardenfors2000conceptual} or induced from cross-linguistic correspondences~\cite{croft2001radical}, but both face scalability challenges. Word embeddings offer a scalable alternative, capturing semantic relations like analogy and similarity~\cite{mikolov2013distributed,vulic2020probing}. However, prior work often relies on static or encoder-based models (e.g., word2vec~\cite{mikolov2013distributed}, BERT~\cite{devlin2019bert}), overlooking decoder-only LLMs and the global structure encoded in their full vocabulary space.


These considerations lead to an intriguing question: what do the semantic networks induced from the \emph{new LLM world} look like? It is “new” in the sense that input embeddings from LLMs remain relatively underexplored, as their next-token prediction objective does not directly encode the meaning of the current token~\cite{liu-etal-2024-fantastic}. It is a “world” because we aim to construct networks over the \emph{entire vocabulary}, offering a global perspective on the internal structure of LLMs. Furthermore, how do such networks differ across LLMs that share the same architecture but vary in computational scale? Given the general trend, established by the scaling law~\cite{kaplan2020scaling}, that larger models tend to perform better, we ask: is this also reflected in the structure of their induced lexico-semantic networks?

We construct a lexico-semantic network from input embeddings of large language models (LLMs), treating vocabulary embeddings as nodes and defining edge weights via similarity metrics. Starting from a fully connected graph, we prune weak edges based on the connectivity hypothesis from semantic map models~\cite{haspelmath2003geometry,liu2024top}, preserving overall connectivity. We analyze the resulting networks across LLMs of different scales, finding small-world structures characterized by high clustering and short path lengths, with larger models exhibiting denser and more complex patterns. To assess practical utility, we examine local subgraphs in three scenarios—common conceptual domains, WordNet-based lexical relations, and a cross-lingual case study—showing strong alignment with human-annotated resources and validating the effectiveness of our approach.


In conclusion, our main contributions are as follows:
\begin{itemize}
    \item We construct a lexico-semantic network using the input embeddings of the entire vocabulary from large language models (LLMs), and show that these networks exhibit the small-world property, indicating strong local clustering and efficient word-to-word connectivity.
    
    \item We conduct a comparative analysis of lexico-semantic networks derived from LLMs of different parameter scales, grounded in the connectivity hypothesis. Our results reveal that larger-scale models display a weaker small-world effect, characterized by longer average path lengths and more complex community structures.
    
    \item We design three evaluation scenarios to analyze the lexico-semantic networks from both global and local perspectives: (1) common conceptual domains, (2) WordNet-based lexical relations, and (3) a cross-lingual case study on evaluative words. In all cases, larger-scale LLMs consistently yield networks with longer semantic paths, reflecting greater relational richness and more nuanced concept structures.
    
\end{itemize}

\section{Related Work}
\label{related work}

\begin{figure*}
    \centering
    \includegraphics[width=0.85\linewidth]{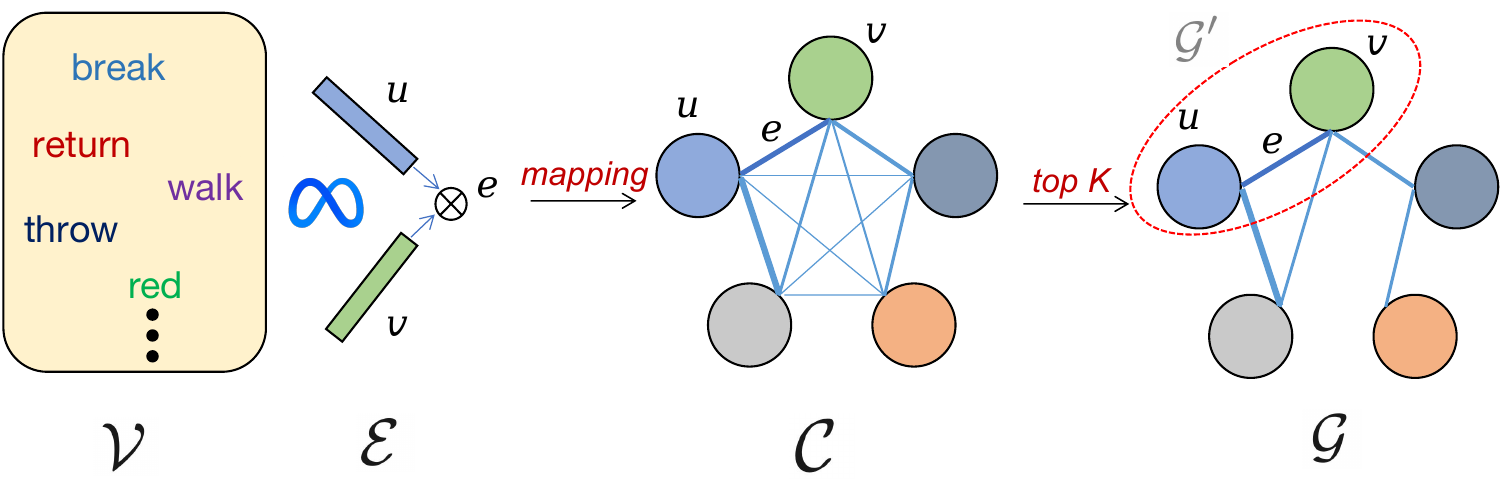}
    \caption{Outline of our lexico-semantic network construction. First, we extract the input word embeddings ($\mathcal{E}$) for the LLM vocabulary ($\mathcal{V}$). Next, we build a complete graph $\mathcal{C}$ by calculating the cosine similarity between all embedding pairs. Finally, we retain edges based on similarity, from highest to lowest, until the graph $\mathcal{G}$ is connected. We then focus on specific connected subgraphs $\mathcal{G'}$ representing certain domains at a local level.}
    \label{fig:intro}
\end{figure*}

\subsection{Word Embedding and Representation}
Contemporary language models represent words or subtokens using continuous vectors based on distributed semantics~\cite{boleda2020distributional}. These high-dimensional vectors can be static~\cite{mikolov2013distributed,Bojanowski2017subword} or context-sensitive~\cite{devlin2019bert}. While effective, they lack the interpretability of semantic feature-based representations~\cite{petersen2023lexical}. Word embeddings exhibit elegant linear relationships~\cite{mikolov2013distributed}, high similarity with human judgments~\cite{vulic2020probing}, and meaningful representations~\cite{Turney2010vector}. Static embeddings are particularly suitable for offline, context-free words, especially monosemous ones. These embeddings are used in the input and output layers of LLMs. Previous research has explored their linear properties~\cite{han2024word}, conceptual space construction~\cite{moullec2025cheaper}, similarity distributions on the tasks of word-pair similarity and analogy~\cite{freestone2024word} and other aspects. They focus on several instances rather than the network structure built by the whole vocabulary. In this paper, we explore the lexico-semantic network via LLM input embeddings, offering a more global and systematic evaluation.

\subsection{Lexico-semantic Network Representations}

A prominent meaning-driven approach to constructing lexico-semantic networks is the \emph{conceptual space framework} (CSF), widely applied in cognitive science~\cite{gardenfors2000conceptual,gardenfors2014geometry,nosofsky1986attention} and neuroscience~\cite{caglar2021conceptual}. CSF models concepts as points or regions in a continuous multidimensional space. For example, the color domain is structured by hue, saturation, and brightness, with color terms corresponding to convex regions~\cite{gardenfors2000conceptual}. Similar structures have been proposed for other domains, such as body parts and kinship~\cite{zwarts2010semantic}.

A data-driven alternative is \emph{semantic map modeling} (SMM), which builds networks from cross-linguistic co-expression patterns found in content words~\cite{guo2012adjectives, petersen2023lexical, dellert2024causal}, function words~\cite{zhang2017semantic}, or constructions~\cite{malchukov2007ditransitive}. Nodes typically represent grammatical or conceptual meanings.

Networks can be built via bottom-up or top-down strategies. The bottom-up method relies on the \emph{connectivity hypothesis}~\cite{haspelmath2003geometry}, linking concepts co-expressed by a single form. The top-down approach~\cite{liu2024top} begins with a complete graph weighted by semantic similarity, then prunes it to retain informative edges.

In this paper, we adopt a data-driven, top-down approach to lexico-semantic network construction, using LLM input embeddings and treating embedding similarity as a proxy for semantic relatedness.

\section{Approach}
\label{sec:approach}


In this section, we first introduce the basic notions and construct a complete graph. We then sparsify the graph based on the revised connectivity hypothesis. Finally, we define global and local metrics to evaluate the lexico-semantic network. The overall pipeline is illustrated in Figure~\ref{fig:intro} and the pseudocode is shown in Algorithm~\ref{alg:network_construction}.

\subsection{Basic Notions}
We define a lexico-semantic network $\mathcal{G} = \{V, E\}$, where $V$ and $E$ are sets of nodes and edges, respectively. Each node $v \in V$ represents a word or a token~\footnote{In LLMs, a token, typically a subword, is the minimal computational unit.}. Each edge $e(u,v) \in E$ connects a pair of nodes $(u,v)$, reflecting their semantic relatedness. If a path $p(u,v)$ exists between nodes $u$ and $v$, they are connected, with path length $L$ defined as the number of edges along the path. If no path exists, $L = \infty$. We use $L$ of the shortest path to indicate the topological distance between two nodes. The network $\mathcal{G}$ is considered connected if every pair of nodes is connected.

A subgraph $\mathcal{G'} = \{V', E'\}$, where $V' \subset V$ and $E' \subset E$, reflects the local topology of $\mathcal{G}$ and typically represents a specific semantic domain, such as adverbs~\cite{zhang2017semantic}, color adjectives~\cite{gardenfors2014geometry}, or qualitative words~\cite{perrin2010polysemous}. Similarly, a subgraph is connected if every pair of nodes has a path. Note that $\mathcal{G}'$ may not connected as a whole.

We define a metric $M$ on $\mathcal{G}$ to measure the relatedness or similarity between nodes. A common metric is cosine similarity
, widely applied in similarity-related tasks. Accordingly, we also utilize the cosine distance $D$ between nodes, which is one minus cosine similarity~\footnote{While cosine distance violates the triangle inequality required by strict distances, we relax this constraint due to its simplicity and widespread use.}.

\subsection{Complete Graph}

We use an LLM to extract input embeddings $\mathcal{E}$ for all tokens in its vocabulary $\mathcal{V}$, treating each token (the minimal computational unit) as a node in the lexico-semantic network. After obtaining the vectorized embeddings, we compute the cosine distance between every pair of nodes to define edge weights. Additionally, we apply centering by subtracting the average vector from each embedding to address anisotropy~\cite{ethayarajh2019contextual}. This results in a complete graph $\mathcal{C}$, where every pair of words is connected.



\subsection{Lexico-semantic Space}
\label{Sec: CS}
We derive a sparsified graph, denoted as the lexico-semantic space $\mathcal{G}$, from the complete graph $\mathcal{C}$. We propose a minimum connectivity hypothesis, which states that $\mathcal{G}$ must remain connected while using the fewest edges possible. The ``connectivity'' ensures that every pair of words is connected, forming a valid space. The ``minimum'' condition favors sparse connections, inspired by the top-down construction of semantic map models, which even use trees (with the least number of edges) to maintain connectivity~\cite{liu2024top}. To achieve this sparsity, we rank edges by weight defined on the metric $M$ and retain the top $K$ ratio of edges until the graph first connects, as higher weights indicate more important connections.

A well-defined $\mathcal{G}$ is also a discrete topological space $\{\mathcal{G}, \mathcal{T}\}$, where $\mathcal{T}$ is the collection of all subsets. We define a subgraph $\mathcal{G'}$ as a subset of $\mathcal{G}$ and it is considered an open set. This is because the intersection and union of any two subgraphs $\mathcal{G_A}$ and $\mathcal{G_B}$ still belong to $\mathcal{T}$:
\begin{equation}
    \small \forall \mathcal{G}_A, \mathcal{G}_B \in \mathcal{T}, \quad  \mathcal{G}_A \cap \mathcal{G}_B \in \mathcal{T}, \quad \mathcal{G}_A \cup \mathcal{G}_B \in \mathcal{T}.
\end{equation}
This is ensured by the ``connectivity'' condition, while a ``minimum'' topological space is required for the conceptual structure.



\subsection{Evaluation}
We evaluate the lexico-semantic network from both global and local perspectives. 

Globally, we compute network statistics to analyze basic properties, connectivity, and small-world characteristics of the spaces built by two models. Small-world characteristics are indicated by a higher clustering coefficient and a shorter shortest path, which are described in detail in Section~\ref{section:GC}.


Locally, we analyze a subgraph \(\mathcal{G'}\) of the conceptual space \(\mathcal{G}\) in three scenarios. Scenario 1 examines common concepts across ten semantic categories, each containing monosemous words, comparing shortest paths within and between groups. Scenario 2 explores shortest-path connections for various WordNet relations. Scenario 3 evaluates a lexico-semantic network of qualitative words, comparing it to the corresponding LLM subgraph. Beyond topology and connectivity, we assess node degree correlations and measure recall and precision against the ground truth.

\section{Experimental Design}
\label{sec:experiment}

\subsection{Large Language Models}
We adopt the Llama series as our LLMs, including Llama2-7B and Llama2-70B~\cite{touvron2023llama}. The dimension of the input embedding is 4096 and 8192, for Llama2-7B and Llama2-70B respectively. 
Also, they share the vocabulary for both models, with the size of vocabulary 32,000. The tokens in the vocabulary are obtained by Byte Pair Encoding~\cite{sennrich2016neural}, merging the bigram with the most frequent co-occurrence. Thus, many tokens are part of a whole word. Besides, tokens at the beginning of a word are different from those in other places, i.e., the end part of a word. For example, 
``man'' in ``policeman'' and ``man''
are different units in the vocabulary. We identify the token appearing the end part of a word by add ``\#'' at the beginning of the token, such as ``\#man''.

\subsection{Scenario 1: Common Concepts}

We construct nine semantic groups representing common concepts, each containing ten frequent words: \textsc{Number}, \textsc{Name}, \textsc{Month}, \textsc{Color}, \textsc{City}, \textsc{Nation}, \textsc{Place}, \textsc{Human}, and \textsc{Furniture}. Some of these groups, such as \textsc{Number}, \textsc{Name}, and \textsc{Month}, naturally exhibit sequential structures. Additionally, we include a \textsc{Random} group as a control. All words in these groups are monosemous and consist of a single token, ensuring that they appear directly in the vocabulary and have well-defined type-level meanings. A full list of the concepts is provided in Table~\ref{tab:concepts_SC} in Appendix~\ref{sec:sce_1}. This scenario investigates how well the embedding-based network captures conceptual similarity, as reflected by the length of shortest paths within and across semantic groups.

\subsection{Scenario 2: WordNet Relations}

In Scenario 2, we examine a subgraph of WordNet~\cite{miller1995wordnet} to assess how structural lexical relations are reflected in the embedding-based network. We use a filtered subset of the public WN18 dataset~\cite{Bordes2013wn18}, comprising 40{,}943 WordNet synsets and 18 relation types. The filtering process follows these criteria: (1) after mapping synsets to their corresponding words, only those present in the LLaMA vocabulary are retained; (2) only the first sense of each synset is kept to represent its prototypical meaning; (3) only relation types involving at least 10 valid word pairs are included; (4) symmetric relation types, such as \textit{hypernym} and \textit{hyponym}, are merged. 

We also introduce two additional wordform-based relation types: \textit{tokenization variant}, which captures differences between tokens with and without a leading underscore (e.g., ``man'' vs. ``\#man''), and \textit{uppercase variant}, which differentiates capitalized from lowercase forms (e.g., ``red'' vs. ``Red''). In total, we include eight relation types, summarized in Table~\ref{tab:relation}. This scenario investigates whether and how these structured lexical relations are preserved in the topology of the embedding-based semantic network.

 
 


\begin{table}[h!]
\centering
\begin{tabular}{ccc}
\toprule
Index & Relation Type & Count \\
\midrule
A & Member of Domain Topic & 51 \\
B & Verb Group & 10 \\
C & Hypernym & 464 \\
D & Has Part & 22 \\
E & Also See & 95 \\
F & Derivationally Related Form & 388 \\
G & Tokenization Variant & 1685 \\
H & Uppercase Variant & 1788 \\
\bottomrule
\end{tabular}
\caption{WordNet Relations and Instance Counts}
\label{tab:relation}
\end{table}

\subsection{Scenario 3: SMM of Qualitative Words}

In Scenario 3, we leverage a cross-linguistic semantic map of adjectives and qualitative terms~\cite{perrin2010polysemous}, which encompasses 22 African languages along with French and English. In this resource, polysemous words within each language are linked to reflect conceptual proximity, forming a semantic map that encodes a universal network of meanings. This map captures cross-linguistic regularities in polysemy patterns, serving as an invariant conceptual structure underlying diverse lexical realizations. The semantic domains covered include dimension, age, value, and color, with concepts represented by capitalized English words (e.g., \textit{BIG}, \textit{SMALL}, \textit{LONG}, \textit{SHORT}, \textit{WIDE}, \textit{DEEP} for the dimension domain).

We filter out words not present in the LLaMA vocabulary, resulting in a final set of 75 concepts from the original 110. The full list of concepts and the human-annotated semantic graph are provided in Appendix~\ref{app:CS}. This scenario evaluates how well the embedding-induced lexico-semantic network aligns with a cross-linguistically grounded conceptual structure.




\section{Results and Analysis}
\label{sec:results}
In this section, we first construct the lexico-semantic network based on the minimum connectivity approach described in Section~\ref{Sec: CS}. We then evaluate the space in three distinct scenarios.

\subsection{Graph Construction}
\label{section:GC}
\begin{figure}
    \centering
    \includegraphics[width=1\linewidth]{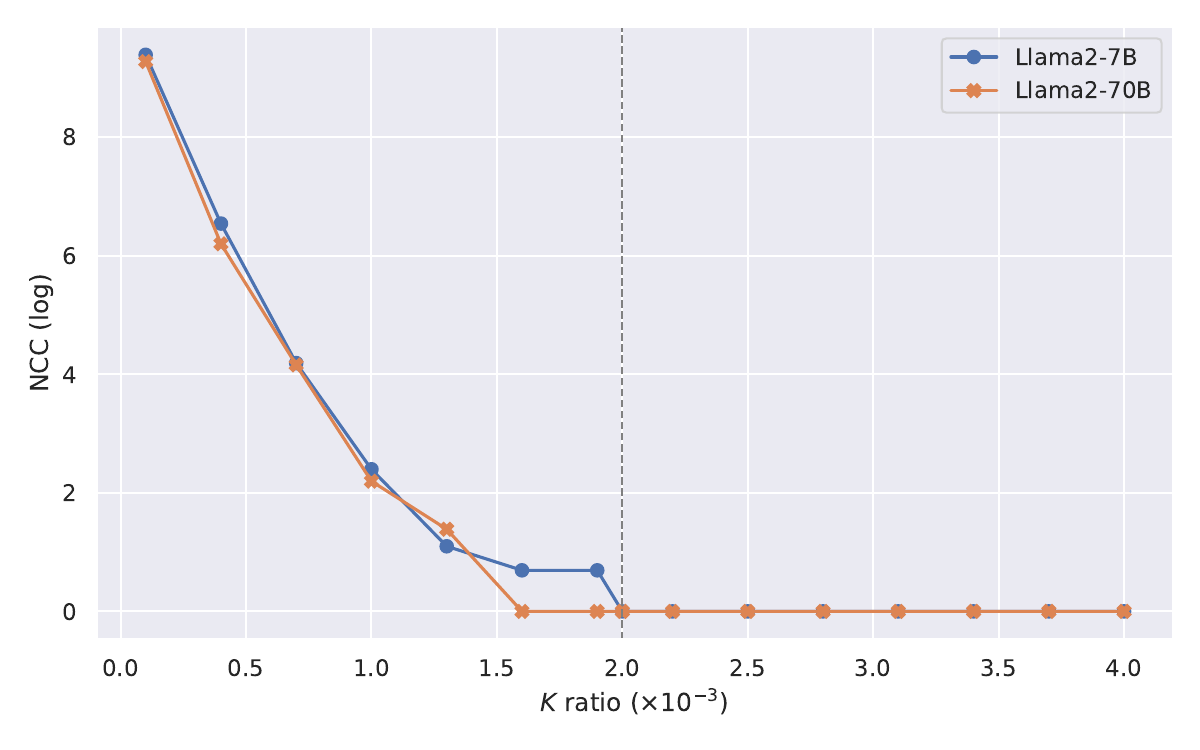}
    \caption{Logarithm of the number of connected components as the top $K$ ratio increases for Llama2-7B and Llama2-70B. A value of zero indicates a fully connected network, while the dotted line marks the first ratio at which both models become connected.}
    \label{fig:K_ratio}
\end{figure}

\paragraph{Choice of $K$.}
To ensure the graph is minimally connected, we extract the top $K$ ratio of edges, where the edge weights are determined by cosine similarity. We incrementally increase the value of $K$ while monitoring the number of connected components (CC), as shown in Figure~\ref{fig:K_ratio}. The graph first becomes connected when the log of the number of CC reaches zero. In our experiments, we selected $K=0.002$, at which point both models become connected. This choice ensures that the same number of edges are used for both models, making the comparison fair.

\begin{table}[h!]
\centering
\begin{tabular}{lcc}
\toprule
\textbf{Statistics} & \textbf{Llama2-7B} & \textbf{Llama2-70B} \\
\midrule
\multicolumn{3}{c}{\textbf{Basic}} \\  
\cmidrule(lr){1-3}
\#Nodes & 32,000 & 32,000 \\
\#Edges & 1,024,000 & 1,024,000 \\
Avg. Degree & 64 & 64 \\
Std. Degree & 68.39 & 58.96 \\
\midrule
\multicolumn{3}{c}{\textbf{Weighted}} \\  
\cmidrule(lr){1-3}
Avg. Degree\_W & 8.76$^{*}$ & 12.23$^{*}$ \\
Std. Degree\_W & 13.78 & 14.02 \\
Threshold & 0.095 & 0.147 \\
\midrule
\multicolumn{3}{c}{\textbf{Small-world}} \\  
\cmidrule(lr){1-3}
GCC ($\uparrow$) & 0.325 & 0.215 \\
ALCC ($\uparrow$) & 0.183$^{*}$ & 0.174$^{*}$ \\
Diameter ($\downarrow$) & 6 & 6 \\
\bottomrule
\end{tabular}
\caption{Statistics for Llama2-7B and Llama2-70B. GCC refers to global clustering coefficient, ALCC to average local clustering coefficient, and Diameter refers to the longest path length. A star ($^*$) indicates high statistical significance between the two models for that metric. Metrics without a star are not tested for significance, typically because they are fixed by design (e.g., \#Nodes, \#Edges) or not applicable for statistical comparison.}

\label{tab:statistics}
\end{table}

\begin{figure*}[h!]
    \centering
    \includegraphics[width=1\linewidth]{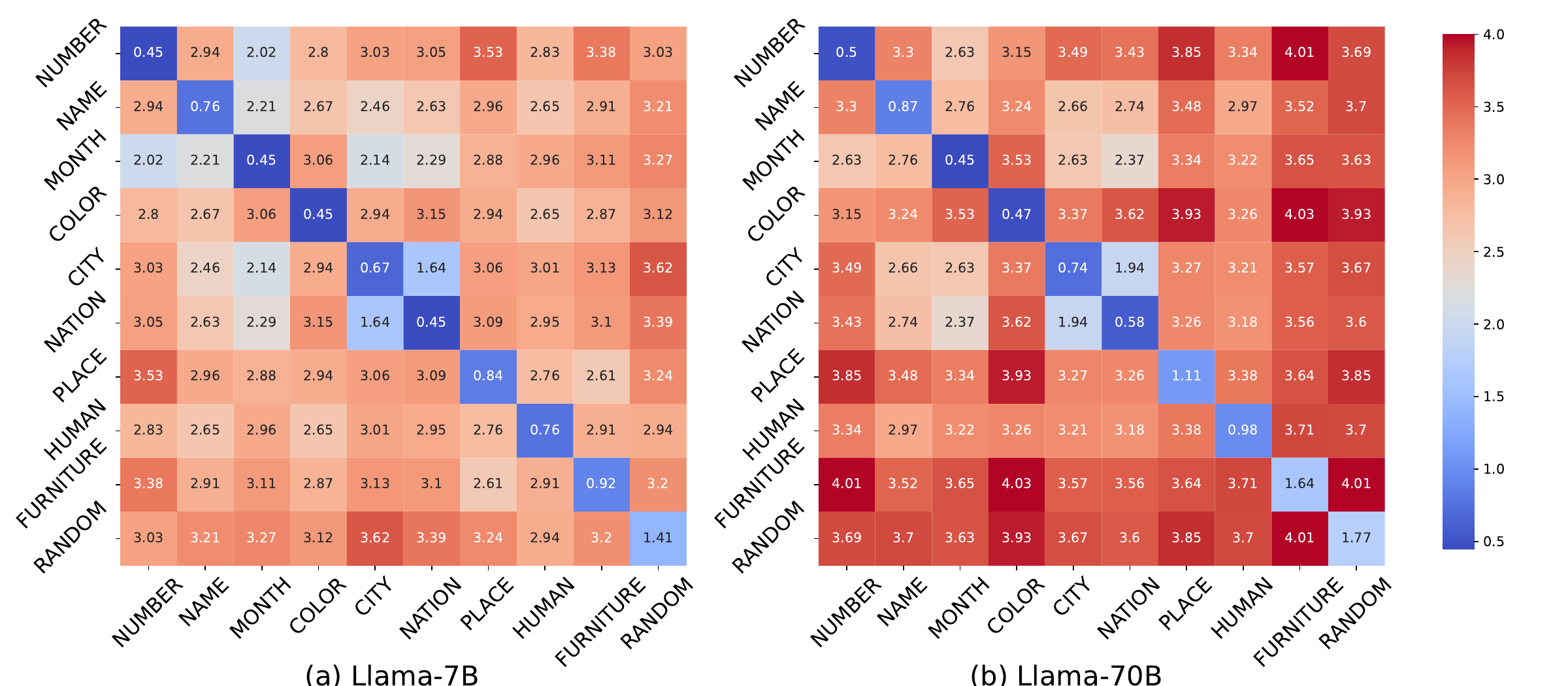}
    \caption{Shortest path lengths among semantic groups for Llama2-7B (left) and Llama2-70B (right).}
    \label{fig:shortest_path}
\end{figure*}

\paragraph{Global Statistics.}
We present the statistics of the lexico-semantic networks for both models in Table~\ref{tab:statistics}, divided into three parts. The \textbf{Basic} section includes the number of nodes (\#Nodes), edges (\#Edges), and the average (Avg. Degree) and standard deviation (Std. Degree) of degrees. The \textbf{Weighted} section calculates the average and standard deviation of weighted degrees (also called Strength or Traffic), along with the equivalent threshold—the minimum weight value in the final graph $G$. The last section focuses on \textbf{Small-world} effects. The global clustering coefficient (GCC) measures graph transitivity—the fraction of actual triangles among all possible triangles in $G$. The local clustering coefficient (ALCC) is calculated as the average of actual connections within neighbors for all nodes. The network diameter is the longest path between any two nodes. Smaller diameters and larger GCC/ALCC values indicate stronger small-world effects. We also calculate the statistical difference of t-test and mark it by a star (*) for a high difference (with a p value less than 0.05).



Llama2-7B and Llama2-70B share similar graph structures, both using the same number of edges. The average degree is 64, but 7B has a flatter degree distribution with a larger standard deviation. In the weighted version, 7B's degree distribution is more concentrated. As shown in Figure~\ref{fig:degree_dist} (Appendix~\ref{app: dd}), 7B exhibits a long-tailed distribution, indicating fewer high-degree (central) nodes.  


Both models exhibit pronounced small-world characteristics, reflected in their high global clustering coefficients (GCC), average local clustering coefficients (ALCC), and short diameters. In contrast, random networks with the same number of edges show substantially lower GCC (0.0032) and ALCC (0.0020). Notably, the observed diameter of 6 corresponds with the Six Degrees of Separation theory~\cite{Milgram1967small}, a phenomenon commonly observed in social networks~\cite{Watts1998collective} and web graphs~\cite{Albert1999internet}. Compared to the smaller model, the 70B model displays a somewhat weaker small-world effect, evidenced by higher GCC and ALCC values, suggesting longer and more complex node interactions. Our subsequent local analyses provide further support for these observations through concrete examples.

\begin{figure}
    \centering
    \includegraphics[width=1\linewidth]{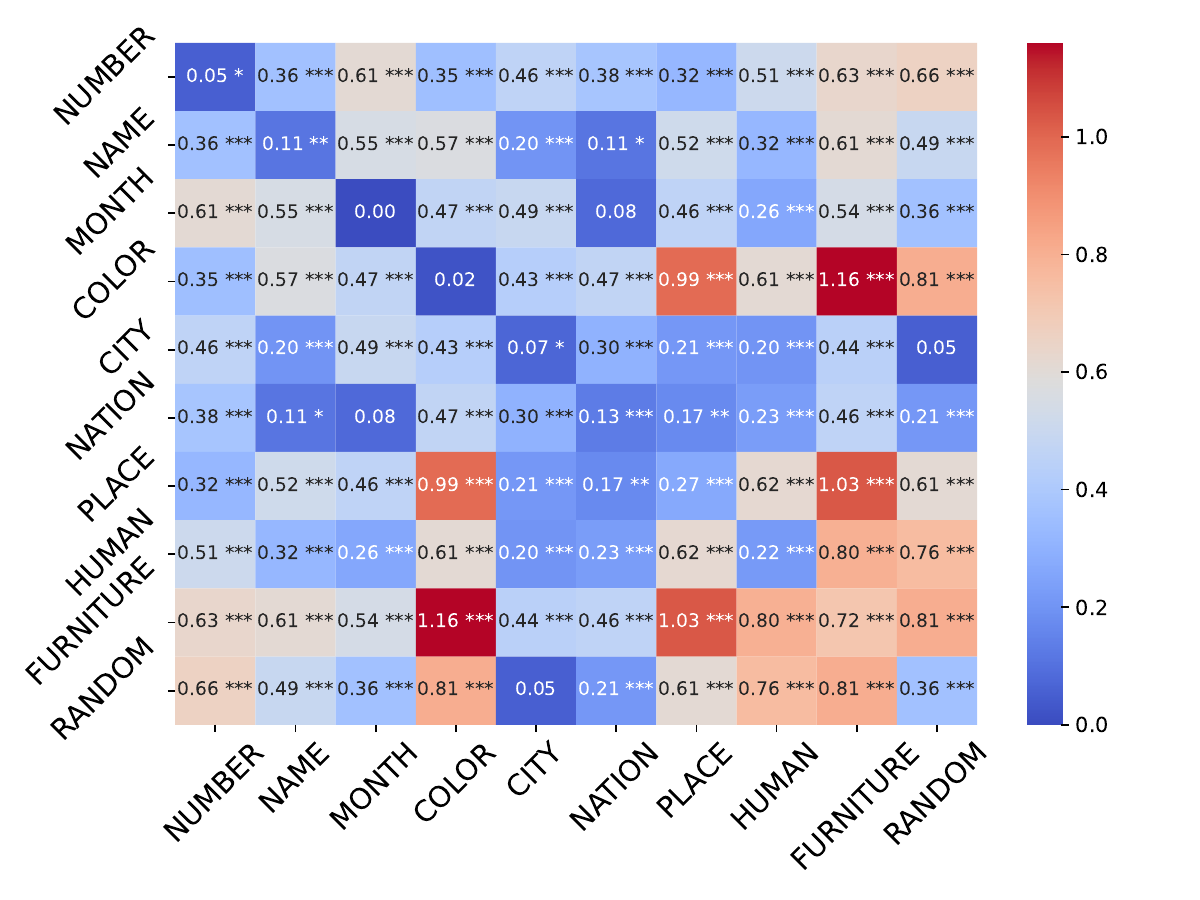}
    \caption{Shortest Path Length Difference (Llama2-70B Minus Llama2-7B) Across Semantic Groups. The number of stars indicate the degree of  the significance level of the difference.}
    \label{fig:difference_map}
\end{figure}

\subsection{Scenario 1}
In this scenario, we evaluate the lexico-semantic networks of both models by calculating the shortest path length between any pair of nodes within and between semantic groups. The shortest path is computed using Dijkstra's algorithm from the NetworkX package~\footnote{\url{https://networkx.org/documentation/stable/reference/algorithms/shortest_paths.html}} with the cosine distance between any two nodes. The average lengths for Llama2-7B and Llama2-70B are shown in Figure~\ref{fig:shortest_path}. Figure~\ref{fig:difference_map} presents the difference heatmap, where the difference is computed as the average length of Llama2-70B minus that of Llama2-7B. The significance of the differences is indicated by the number of stars (one, two, or three), corresponding to p-values of 0.05, 0.01, and 0.001 in the t-test, respectively.

\begin{figure}
    \centering
    \includegraphics[width=1.0\linewidth]{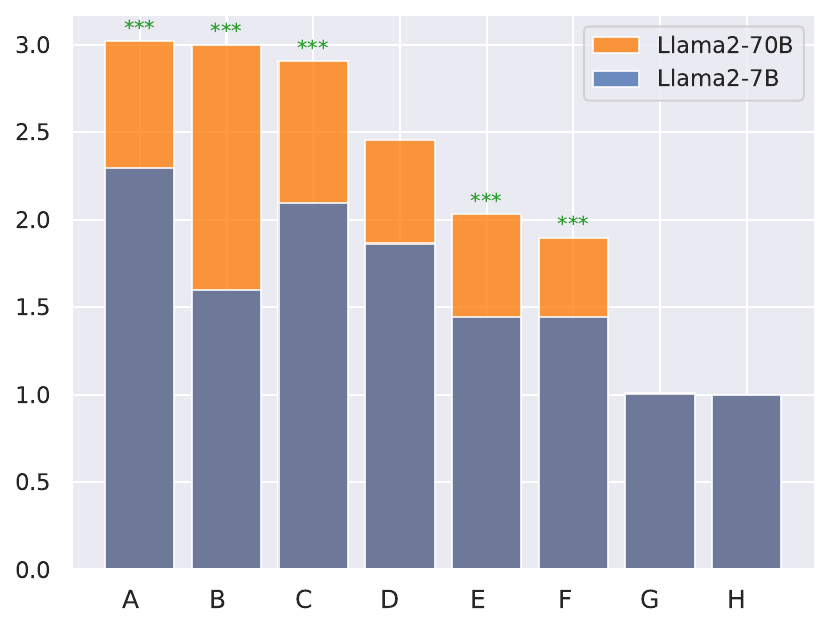}
    \caption{Averaged shortest path length across relation types for both models. The number of stars indicates the significance of the difference.}
    \label{fig:wordnet_path}
\end{figure}

\begin{figure*}[h!]
    \centering
    \begin{subfigure}[t]{0.48\linewidth}
        \centering
        \includegraphics[width=\linewidth]{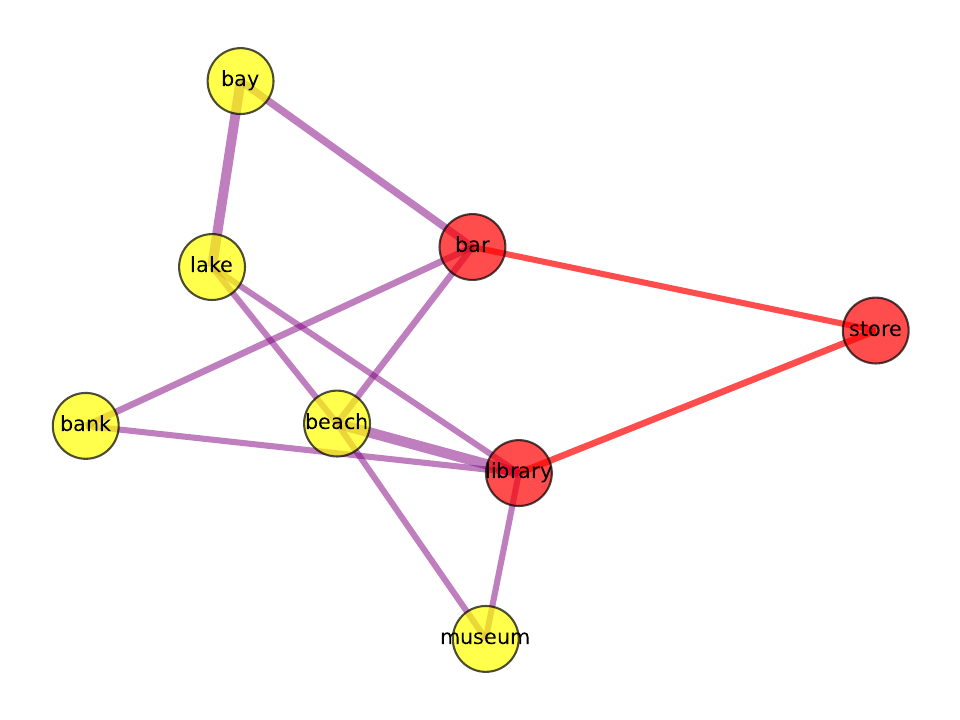}
        \caption{Llama2-7B}
        \label{fig:example_7B}
    \end{subfigure}
    \hfill
    \begin{subfigure}[t]{0.48\linewidth}
        \centering
        \includegraphics[width=\linewidth]{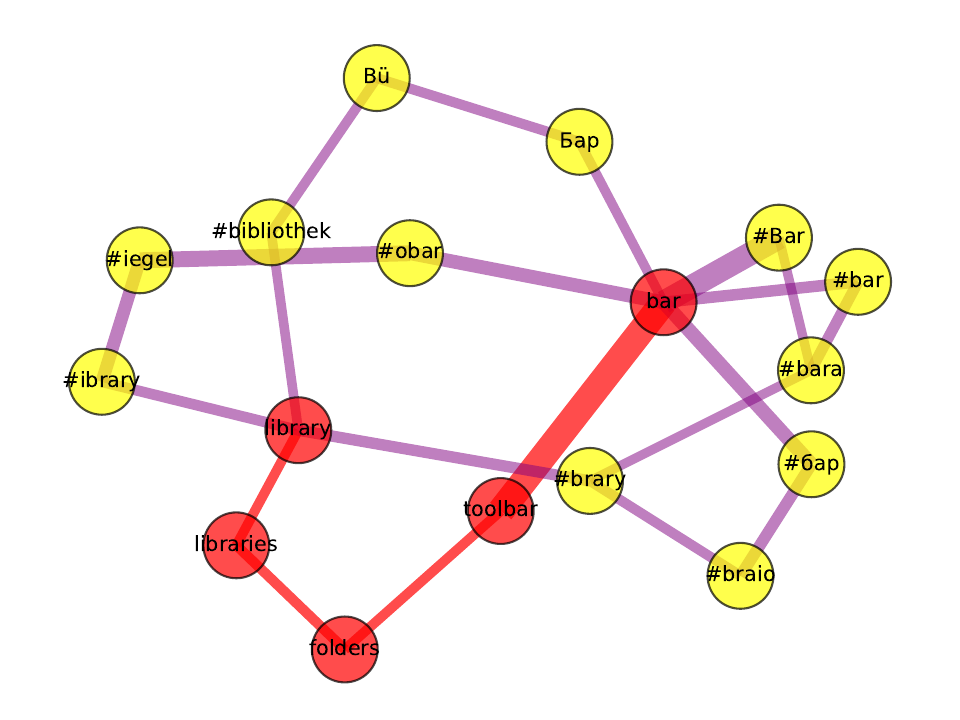}
        \caption{Llama2-70B}
        \label{fig:example_70B}
    \end{subfigure}
    \caption{Shortest paths between ``bar'' and ``library'' in Llama2-7B and Llama2-70B. Edge width indicates weight; \textcolor{red}{red} highlights mark the shortest path. The 70B model shows a more complex structure involving similar forms, multilingual links, and disambiguation clusters.}
    \label{fig:example_paths}
\end{figure*}

Our results indicate that Llama2-70B generally exhibits longer path lengths than Llama2-7B, both within and across semantic groups, suggesting that Llama2-70B uncovers relationships through more complex pathways. Figures~\ref{fig:example_7B} and~\ref{fig:example_70B} illustrate the six shortest paths between “bar” and “library.” The shortest path (highlighted in red) in Llama2-70B shows greater variation in word forms and conjugations. Moreover, alternative paths involve multilingual links and diverse connections that assist in disambiguation—for instance, differentiating the building “library” from its software-related sense. This richer semantic representation may account for the superior performance of larger-scale models.



\begin{figure*}[h!]
    \centering
    \begin{subfigure}[t]{0.48\linewidth}
        \centering
        \includegraphics[width=\linewidth]{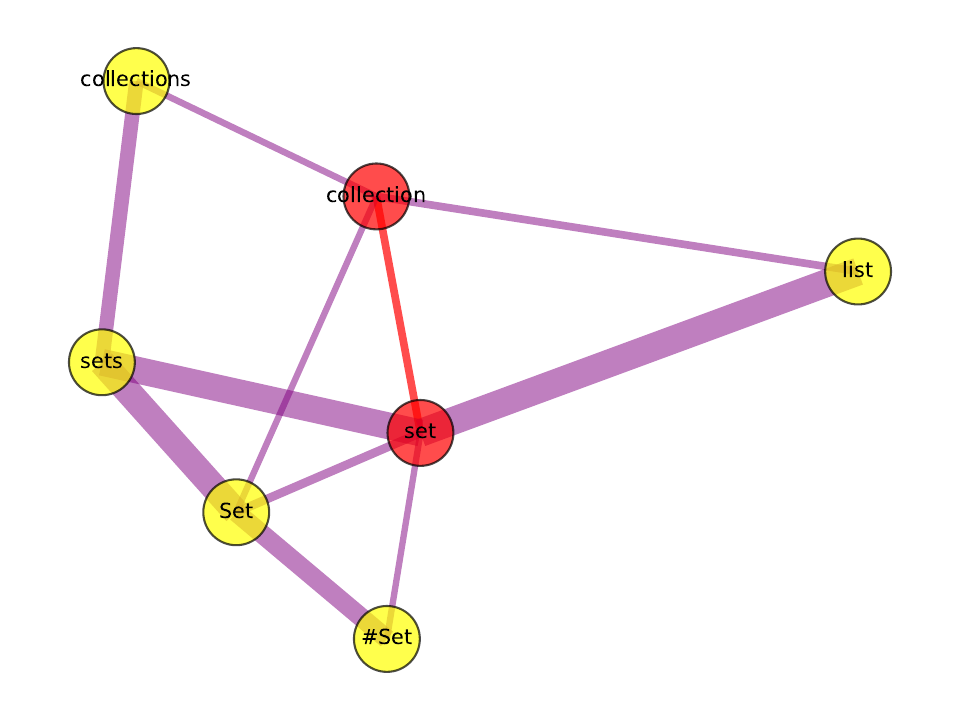}
        \caption{Llama2-7B}
        \label{fig:example_hyp_7B}
    \end{subfigure}
    \hfill
    \begin{subfigure}[t]{0.48\linewidth}
        \centering
        \includegraphics[width=\linewidth]{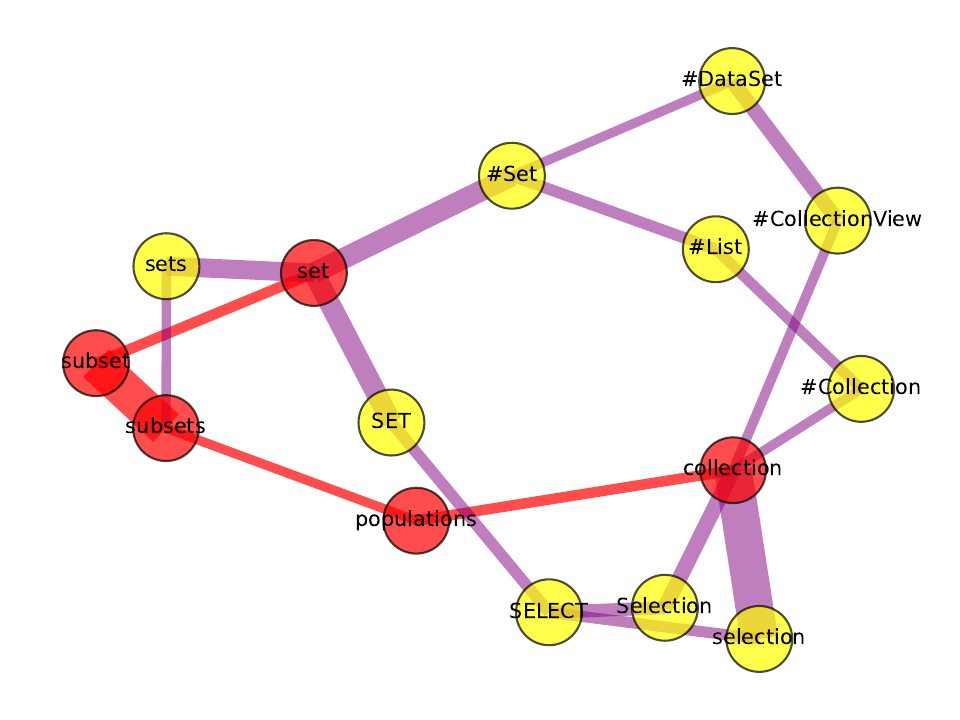}
        \caption{Llama2-70B}
        \label{fig:example_hyp_70B}
    \end{subfigure}
    \caption{Shortest paths between ``collection'' and ``set'' in Llama2 models. \textcolor{red}{Red} highlights indicate the shortest path. The 70B model shows a more logical semantic transition from singular to plural concepts.}
    \label{fig:example_hyp_paths}
\end{figure*}

When analyzing per group, both models show shorter paths within the same group, and the trends are similar for different group pairs. Among the groups, COLOR shows the shortest paths, while FURNITURE exhibits the longest. This may be due to polysemy in FURNITURE, where terms like ``chair'' refer to both furniture and a human (e.g., a person referred to as a ``chair''). This is further supported by the longer paths when comparing FURNITURE to other groups. Conversely, NATION and CITY tend to have shorter paths.

\subsection{Scenario 2}
In the second scenario, we evaluate the network for word pairs from different relation types, as shown in Table~\ref{tab:relation}. For each pair, we compute the shortest path and display the length for both models. The significance of the differences is indicated by the number of stars. The results are shown in Figure~\ref{fig:wordnet_path}.


The results mirror those of Scenario 1: Llama2-70B generally exhibits longer paths than Llama2-7B. For most relation types, the path lengths in 70B are significantly greater, except for word form–related relations where both models maintain direct edges between word pairs. Longer paths are observed in relations such as \textit{member of domain topic}, \textit{verb group}, and \textit{hypernym}, reflecting more complex connection chains. For instance, in the \textit{hypernym} relation between “set” and “collection,” Llama2-70B demonstrates a more nuanced path, gradually linking the singular “set” to the plural “collection” with conjugation correlations, as illustrated in Figures~\ref{fig:example_hyp_7B} and~\ref{fig:example_hyp_70B}. This suggests that larger-scale models capture more logical and intricate conceptual relations.




\subsection{Scenario 3}
In Scenario 3, we construct a lexico-semantic network for qualitative words, with a reference (GT) from cross-lingual research. The statistics for the spaces built by Llama2-7B (7B), Llama2-70B (70B), and GT are shown in Table~\ref{tab:SMM_stats}.

\begin{table}[h]
\centering
\begin{tabular}{lccc}
\toprule
\textbf{Statistics} & \textbf{7B} & \textbf{70B} & \textbf{GT} \\
\midrule
\multicolumn{4}{c}{\textbf{Basic}} \\  
\cmidrule(lr){1-4}
\#Nodes & 75 & 75 & 75 \\
\#Edges & 293 & 130 & 37 \\
Avg. Degree & 7.813 & 3.467 & 0.987 \\
Std. Degree & 5.724 & 3.021 & 0.959 \\
\midrule
\multicolumn{4}{c}{\textbf{Weighted}} \\  
\cmidrule(lr){1-4}
Avg. Degree\_W & 0.963 & 0.611  & - \\
Std. Degree\_W & 0.736 & 0.546 & - \\
Avg. Weight & 0.123 & 0.176 & - \\
\midrule
\multicolumn{4}{c}{\textbf{Connectivity}} \\  
\cmidrule(lr){1-4}
\#Component ($\downarrow$) & 8 & 17 & 39 \\
\#Single ($\downarrow$) &7  & 16 & 26 \\
\midrule
\multicolumn{4}{c}{\textbf{Reference with GT}} \\
\cmidrule(lr){1-4}
Correlation ($\uparrow$) & 0.466 & 0.449 & 1 \\
Recall ($\uparrow$) & 0.568 & 0.378 & - \\
Precision ($\uparrow$) & 0.072 & 0.108 & - \\
\bottomrule
\end{tabular}
\caption{Statistics for Llama2-7B, Llama2-70B, and GT across four dimensions. ``\#Component'' represents the number of connected components, and ``\#Single'' indicates the number of nodes with zero degree. In the bottom section, we report degree correlation, recall, and precision.}
\label{tab:SMM_stats}
\end{table}


Overall, the model-constructed networks contain more edges and fewer connected components and isolated nodes than those built by human experts. Expert-constructed graphs are based on corpus co-occurrence across at least three languages, a process especially difficult for low-resource languages. In contrast, model embeddings produce much denser graphs. Notably, Llama2-70B yields a sparser network than Llama2-7B. Compared to the ground truth (GT), the automatic networks show moderate correlation and coverage (recall), indicating that the lexico-semantic network partially captures universal cross-linguistic patterns of semantic relatedness in polysemy. However, precision is lower due to the larger number of edges, suggesting embeddings can serve as a preliminary space for linguists to further refine.

\section{Conclusion}
This paper investigates the construction of lexico-semantic networks using input embeddings from large language models (LLMs). We analyze and compare the network properties of LLMs with different scales across three scenarios. Our findings show that the lexico-semantic network can be effectively constructed from embeddings, which exhibit a small-world clustering effect. Additionally, models with more parameters tend to explore longer and more complex paths between concepts, partially supporting the ``scaling law''~\cite{kaplan2020scaling}. This study also provides an efficient approach to constructing conceptual spaces, potentially benefiting fields such as language typology and cognitive science.

\section{Limitations}
We acknowledge several limitations in our work. First, we mainly concentrate on the monosemous words. However, words can be ambiguous, particularly for homonyms that encompass multiple unrelated meanings. Second, our evaluation is limited to two models, Llama2-7B and Llama2-70B. Results may differ with models of different architectures or parameter scales. Additionally, we focus only on input embeddings and do not explore the properties of output embeddings, which may also capture individual word representations. Finally, the length of the shortest path in the lexico-semantic network is not a definitive metric for embedding quality, and we plan to explore more sophisticated metrics to better reflect the characteristics of these spaces.

\section{Ethics Statement}
We do not foresee immediate ethical concerns arising from our research. However, there may be unintended biases in the connections between concepts, such as those involving gender and job ranks. These biases may stem from biased embeddings~\cite{Bordes2013wn18, bolukbasi2016man}.



\bibliography{custom}
\bibliographystyle{acl_natbib}
\clearpage

\appendix

\section{Appendix}
\label{sec:appendix}

\subsection{Construction of Lexico-Semantic Network}
\label{sec:appendix_network_building}
Algorithm~\ref{alg:network_construction} illustrates the procedure for constructing the lexico-semantic network. The process involves extracting token embeddings from a language model, computing pairwise cosine distances, and retaining a subset of edges to ensure minimal connectivity. This algorithm corresponds to the method described in Section~\ref{Sec: CS}.

\begin{algorithm*}
\caption{Constructing Lexico-Semantic Network with Minimal Connectivity}
\label{alg:network_construction}
\begin{algorithmic} 
\Require Token set $\mathcal{V}$, LLM model $M$
\Ensure Lexico-semantic graph $\mathcal{G} = (V, E)$ and number of selected edges $K$
\State $V \gets \mathcal{V}$
\State Extract embeddings $\mathcal{E}$ from $M$ for each $v \in V$
\State Center each embedding: $\mathcal{E} \gets \mathcal{E} - \text{mean}(\mathcal{E})$
\State Compute pairwise cosine distances $\text{D}(u,v)$ for all $u,v \in V$
\State Initialize empty graph $\mathcal{G} = (V, \emptyset)$
\State Sort all edges $(u,v)$ by increasing $\text{D}(u,v)$ to get edge list $L$
\For{each edge $(u,v)$ in $L$}
    \State Add edge $(u,v)$ to $E$
    \If{$\mathcal{G}$ is connected}
        \State \Return $\mathcal{G},\ K = |E|$
    \EndIf
\EndFor
\end{algorithmic}
\end{algorithm*}

\subsection{Common Concepts in Scenario 1}
\label{sec:sce_1}
Table~\ref{tab:concepts_SC} lists specific concepts from different semantic groups, with each group containing ten common concepts.

\begin{table*}[h!]
    \centering
    \begin{tabular}{cc}
        \toprule
        \textbf{Semantic Group} & \textbf{Words} \\
        \midrule
        NUMBER & one, two, three, four, five, six, seven, eight, nine, ten \\
        NAME & Alice, Bob, Carol, Dave, Francis, Grace, Hans, Ivan, Zach, Mike \\
        MONTH & January, February, March, April, May, June, July, August, September, October \\
        COLOR & red, orange, yellow, green, blue, brown, black, white, grey, gray \\
        CITY & Taiwan, York, Cambridge, Oxford, Berlin, Paris, Washington, Rome, Tokyo, Toronto \\
        NATION & China, America, England, UK, Germany, France, USA, Italy, Japan, Spain \\
        PLACE & factory, concert, museum, library, bar, zoo, park, theater, hospital, church \\
        HUMAN & female, male, man, woman, human, boy, girl, elder, gentleman, guys \\
        FURNITURE & chair, desk, table, bed, cabinet, computer, lamp, mirror, house, room \\
        RANDOM & conscious, distance, measure, almost, paste, sun, friend, other, waste, tongue \\
        \bottomrule
    \end{tabular}
    \caption{Specific words from different semantic groups.}
    \label{tab:concepts_SC}
\end{table*}

\subsection{Conceptual Spaces in Scenario 3}
\label{app:CS}
\begin{figure*}
    \centering
    \includegraphics[width=1.1\linewidth]{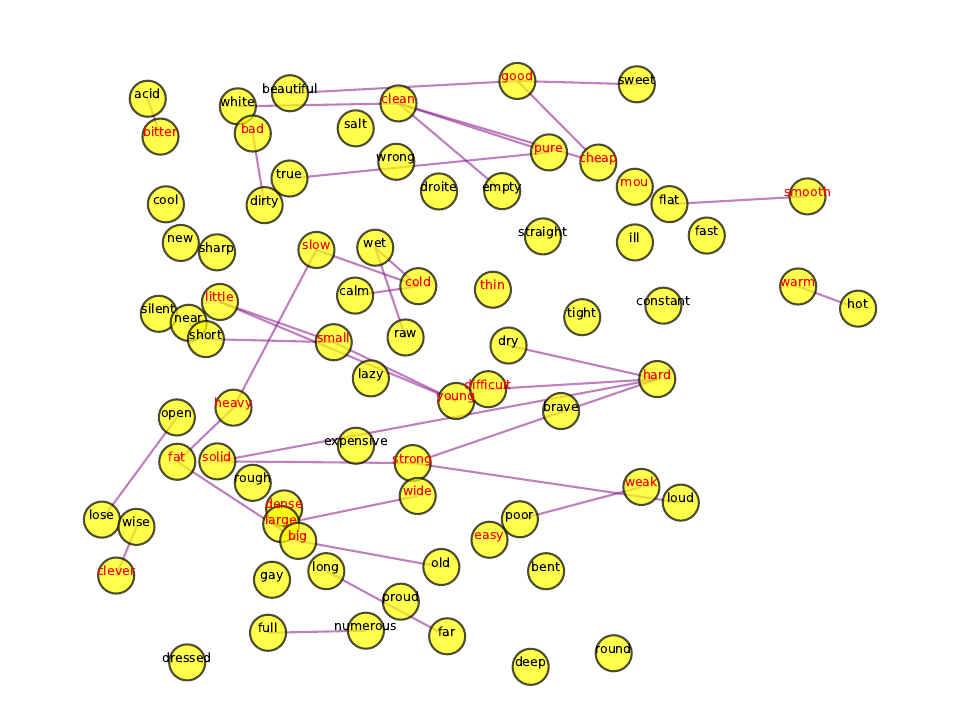}
    \caption{A semantic map for the domain of qualitative words, with federative notions which have a higher degree highlighted in red.}
    \label{fig:SMM_human}
\end{figure*}

Scenario 3 presents a human-annotated conceptual space for qualitative words, as shown in Figure~\ref{fig:SMM_human}. Each concept is represented by an English word. Nodes are connected if a pair of concepts co-occur as a polysemous word in at least three languages. Nodes marked in red represent federative words, indicating a shared concept with a higher degree.

\subsection{Degree Distribution}
\label{app: dd}
\begin{figure*}
    \centering
    \includegraphics[width=1.0\linewidth]{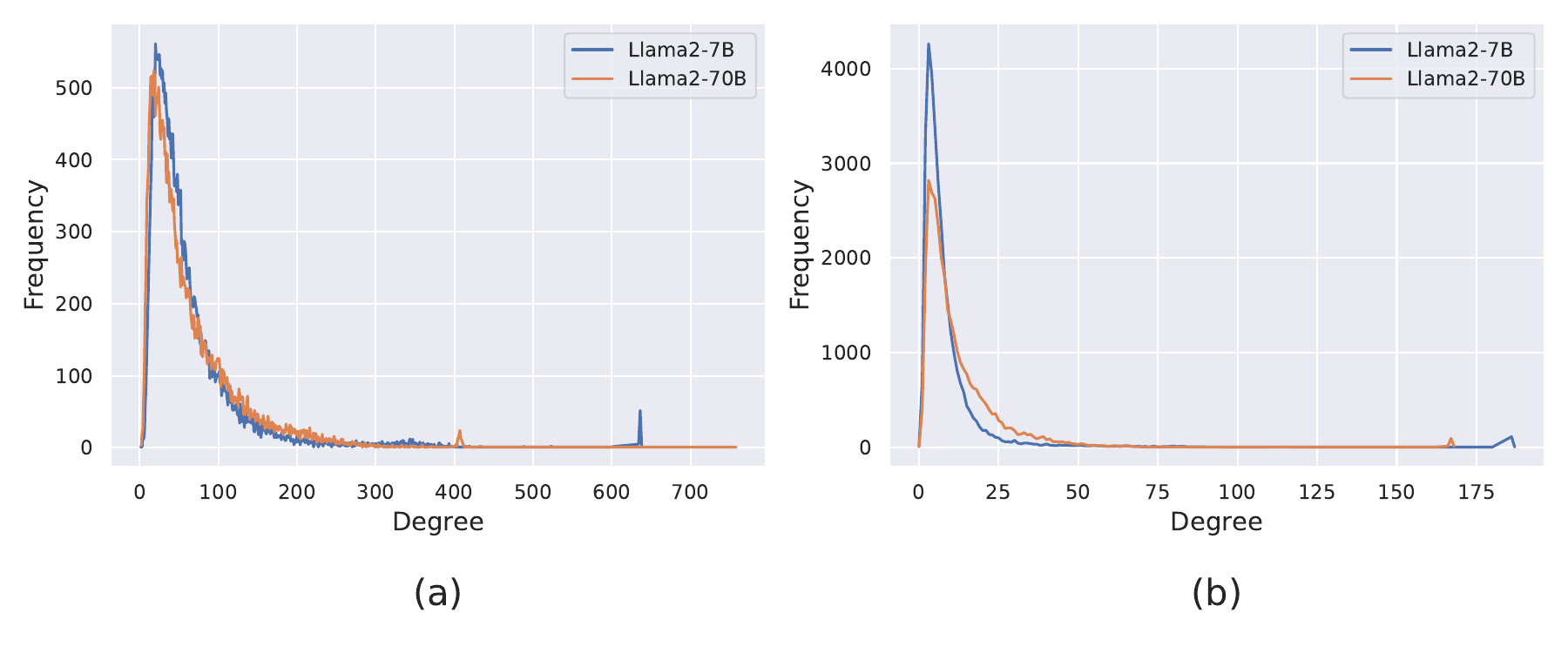}
    \caption{Comparison of the degree distribution for both models: (a) unweighted degree and (b) weighted degree.}
    \label{fig:degree_dist}
\end{figure*}

We show the distribution of node degrees for spaces generated by two models, Llama2-7B and Llama2-70B. Figure~\ref{fig:degree_dist}(a) displays the unweighted degree distribution, while (b) shows the weighted distribution. The results indicate that the 70B model has a less pronounced long-tail distribution, with more nodes having relatively larger degrees.

\end{document}